%% file: root.tex
\documentclass[letterpaper, 10pt, conference]{ieeeconf}      

\IEEEoverridecommandlockouts                              

\usepackage{graphics} 
\usepackage{epsfig}   
\usepackage{mathptmx} 
\usepackage{times}    
\usepackage{amsmath}  
\usepackage{amssymb}  

\usepackage{graphicx}
\usepackage{subcaption}
\usepackage{gensymb}
\usepackage{algorithm}
\usepackage[noend]{algpseudocode}
\usepackage{array}
\usepackage{multirow}
\usepackage{multicol}
\usepackage{xcolor}

\title{\LARGE \bf
    DenseCAvoid: Real-time Navigation in Dense Crowds using Anticipatory Behaviors
}

\author{Adarsh Jagan Sathyamoorthy$^{1}$, Jing Liang$^{1}$, Utsav Patel,  Tianrui Guan, Rohan Chandra, and Dinesh Manocha
\thanks{$^{1}$ Authors contributed equally. 
All authors are with the University of Maryland, College Park.
        }%
}

\input{macros.tex}

\begin{document}
\maketitle
\thispagestyle{empty}
\pagestyle{empty}

\input{Sections/0_Abstract.tex}
\input{Sections/1_Introduction.tex}
\input{Sections/2_Previous_Work.tex}

\input{Sections/3_Background.tex}
\input{Sections/4_OurMethod.tex}
\input{Sections/5_Results_and_Evaluations.tex}

\input{Sections/6_Conclusion.tex}

\bibliography{references} 
\bibliographystyle{ieee}

\end{document}

%% file: macros.tex
\usepackage{xcolor}


%% file: Sections/0_Abstract.tex
\begin{abstract}
We present DenseCAvoid, a novel navigation algorithm for navigating a robot through dense crowds and avoiding collisions by anticipating pedestrian behaviors. Our formulation uses visual sensors and a pedestrian trajectory prediction algorithm to track pedestrians in a set of input frames and provide bounding boxes that extrapolate the pedestrian positions in a future time. Our hybrid approach combines this trajectory prediction with a Deep Reinforcement Learning-based collision avoidance method to train a policy to generate smoother, safer, and more robust trajectories during run-time.  We train our policy in realistic 3-D simulations of static and dynamic scenarios with multiple pedestrians. In practice, our hybrid approach generalizes well to unseen, real-world scenarios and can navigate a robot through dense crowds ($\sim$1-2 humans per square meter) in indoor scenarios, including narrow corridors and lobbies. As compared to cases where prediction was not used, we observe that our method reduces the occurrence of the robot freezing in a crowd by up to 48\%, and performs comparably with respect to trajectory lengths and mean arrival times to goal.



\end{abstract}

%% file: Sections/1_Introduction.tex
\section{INTRODUCTION}
Mobile robots are increasingly being used in different scenarios that are crowded with pedestrians and other obstacles. For instance, robots are being used for room service in hotels, package and food delivery, or as caretakers in hospitals. Furthermore, they are used for surveillance in public places such as malls, airports, etc. These environments can be crowded with high pedestrian density (e.g., 1-2 pedestrians per square meter). In such scenarios, a robust and efficient collision avoidance method is crucial to ensure the safety of the robot and the humans and for better functionality.

The problem of robot navigation among dynamic obstacles has been well studied in robotics and related areas. There is a large body of work on {\em classic navigation techniques} based on potential fields, velocity obstacles, and dynamic windows~\cite{RVO,ORCA,NH-ORCA,Lav06,FoxThurn}. Recently, many collision avoidance methods based on machine learning have been proposed~\cite{JHow1,JiaPan1,End2End,WB1,Alahi,Perception2Decision} and have shown considerable promise in real-world scenarios. These {\em learning-based methods} can be directly integrated with current 2-D or 3-D lidars or cameras and do not require accurate sensing of the states of the obstacles. 

In practice, dense crowds pose several challenges for robot collision avoidance. First, pedestrian motions in such crowds can be highly unpredictable and non-smooth~\cite{ReachBera}. Second, the robot must be able to react to sudden changes in pedestrian motion to avoid collisions. Current learning-based collision avoidance methods \cite{JHow1,WB1,JHow2} work well for sparse or moderately dense crowds, but either result in collisions or oscillations \cite{JiaPan1} as the crowd density increases. Another common phenomenon in such cases is the freezing robot problem \cite{Unfrozen,freeze, freeze2}, where the navigation method completely halts the robot, considering all forward velocities as unsafe. Moreover, if the pedestrians obstructing the robot do not move to give way, the robot could stall indefinitely. 


\begin{figure}[t]
      \centering
      \includegraphics[width = \columnwidth]{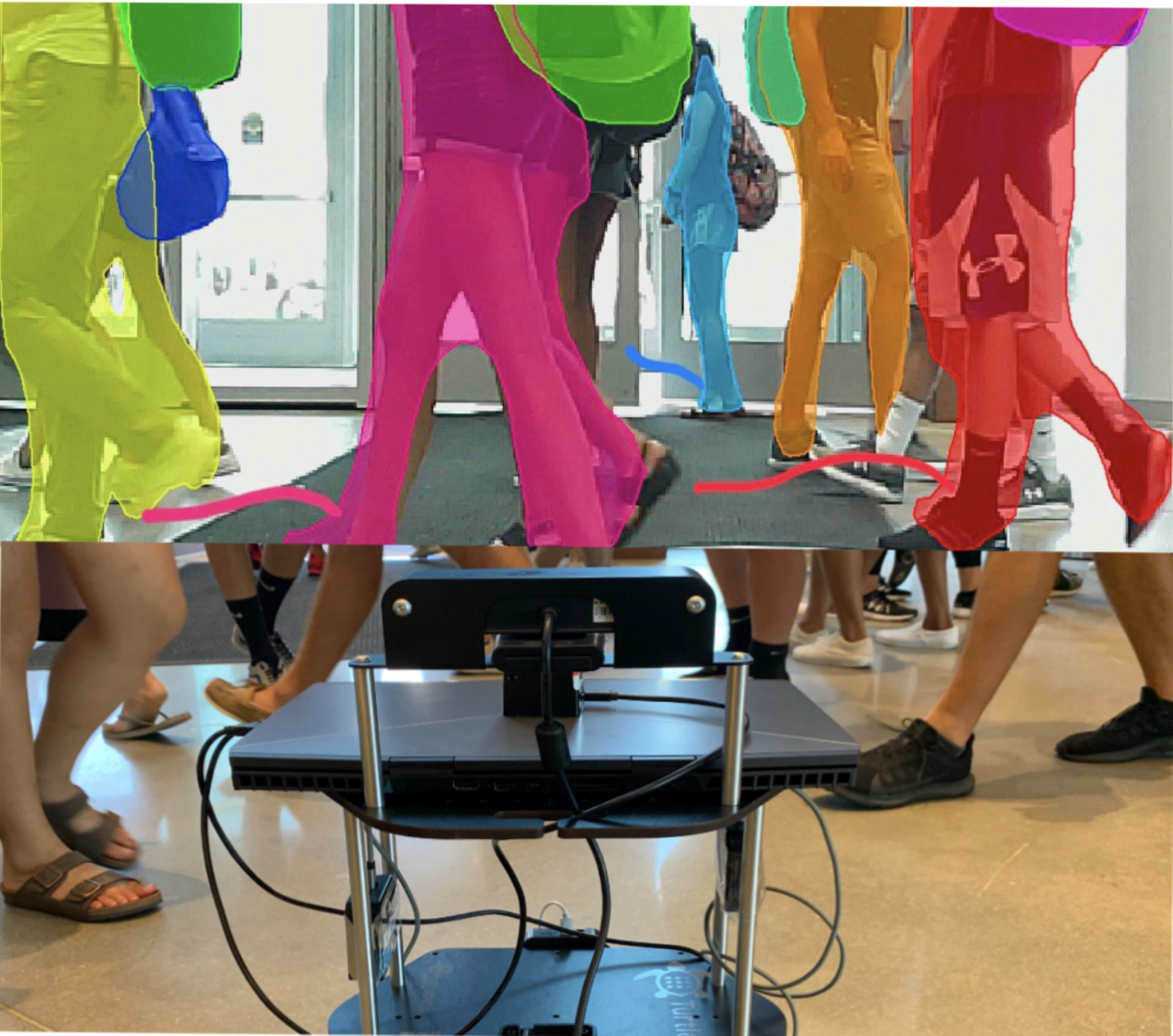}
      \caption {We explicitly track and predict pedestrian motions (marked for the three pedestrians above) and use it to train a Deep Reinforcement learning-based collision avoidance policy. Our network implicitly learns to reduce the occurrence of the freezing robot problem and generates smooth robot trajectories in dense crowds.}
      \label{Cover}
    \vspace{-15pt}
   \end{figure}

\noindent {\bf Main Results:} We present a novel algorithm (DenseCAvoid) for safe robot navigation in dense crowds. Our approach is hybrid and combines techniques based on Deep Reinforcement Learning (DRL) with navigation methods that use pedestrian trajectory prediction. As a result, our approach provides the benefits of learning-based methods in terms of handling noisy sensor data, along with the benefits of a navigation method that explicitly predicts the trajectory and behavior of each pedestrian in an anticipatory manner. The latter enables our hybrid approach to robustly deal with new or unforeseen scenarios, which are quite different from the synthetic training data.

We present a new deep reinforcement learning algorithm, which includes a modified network, reward function and training scenarios that takes these characteristics into account. We  use a state-of-the-art pedestrian trajectory prediction method~\cite{chandra2019robusttp} that can handle dense scenarios. Our collision avoidance policy uses these explicitly predicted pedestrian motions to train, using a policy gradient method known as Proximal Policy Optimization (PPO)\cite{PPO}. During run-time, our method uses raw sensor data from a lidar, a depth camera, the robot's odometry, and pedestrian prediction to generate smooth, collision-free trajectories. Our main contributions include:


\begin{itemize}
\item A new end-to-end DRL-based collision avoidance policy that is combined with a human motion prediction algorithm to anticipate pedestrian motion and generate smooth trajectories in dense crowds. This results in an increase by up to 74\% in rates of reaching the goal when compared to times when prediction was not used.

\item A motion prediction algorithm that is general and can handle dense scenarios, that is more robust than a simple linear motion model.

\item A novel network structure and reward function that take multiple sensor inputs and pedestrian prediction data to train a policy using PPO. This results in a reduction by up to 48\% in the occurrence of the freezing robot problem.

\item Complex 3-D simulations of indoor environments with pedestrians and static obstacles for training and bench-marking DRL methods. 
\end{itemize}

%% file: Sections/2_Previous_Work.tex
\section{PRIOR WORK AND BACKGROUND}
In this section, we briefly cover prior work in pedestrian tracking, motion prediction and learning-based collision avoidance methods.

\subsection{Pedestrian Tracking and Motion Models}
Object and pedestrian detection has been widely studied in computer vision. Some of the most accurate methods are based on deep learning including R-CNN \cite{2013arXiv1311.2524G} and its faster variants \cite{fastrcnn,fasterrcnn,maskrcnn}, which use a selective search area to optimize the object detection problem. Other works in learning-based tracking include \cite{rtdl1,rtdl2,rtdl4,rtdl5-online153}. Many of these, learning-based methods lack real-time performance that is needed for navigation in dense environments. Moreover, highly accurate methods such as \cite{deepsort} and \cite{rt1} require high-quality detection features for reliable performance. 

Several motion models have been used to improve pedestrian tracking accuracy \cite{cem-lin2,mht-lin1,edmt-lin3,lfnf-lin4}. However, most of these methods assume a constant linear velocity or acceleration models for the pedestrians. However, these methods cannot characterize pedestrian dynamics accurately in dense settings~\cite{bera2014realtime}. Non-linear motion models such as RVO~\cite{RVO} and its variants have been shown to work well for tracking in dense crowd videos. Social Force model~\cite{bera2014adapt}, LTA~\cite{5459260}, and ATTR~\cite{yamaguchi2011you} are other non-linear motion models that have been used for pedestrian tracking in low to medium density crowds.

\textbf{YOLOv3}: In our method, we use YOLOv3~\cite{redmon2018yolov3}, a real-time high accuracy variant of YOLO~\cite{redmon2016you} for pedestrian detection and tracking in dense crowd. Given an RGB or a depth image and an object of interest (pedestrians, in our case), it outputs the bounding box coordinates over all the detected objects in the image. It can be combined with front-based RVOs to further improve the accuracy~\cite{chandra2019robusttp}.


\begin{figure}[t]
      \centering
      \includegraphics[width = \columnwidth]{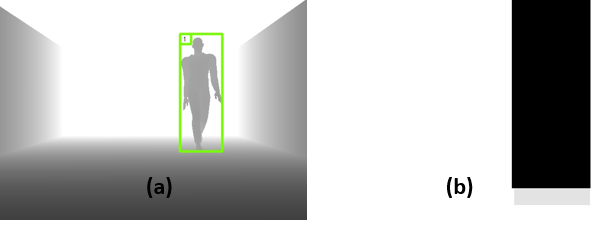}
      \caption {(a) Tracking a moving pedestrian in a depth image. (b) Prediction output for a future time step. The white space in the output is the admissible free space, and the black bounding box denotes the space the pedestrian would occupy in the future.}
      \label{fig:TrackPredict}
    \vspace{-15pt}
  \end{figure}

\subsection{Pedestrian Trajectory Prediction and Navigation}
There has been extensive research in predicting object or pedestrian trajectories in computer vision and robotics. Early works include formulations such as Bayesian~\cite{early1-bayesian}, Monte Carlo simulation~\cite{early5}, Hidden Markov Models~\cite{early2-HMM}, and Kalman Filters \cite{early4-kalman}. Deep learning-based prediction methods mostly utilize Recurrent Neural Networks (RNNs)~\cite{rnn1} and Long Short-Term Memory (LSTM). Hybrid methods using a combination of RNNs and other deep learning architectures such as Convolutional Neural Networks (CNNs), Generative Adversarial Networks (GANs) and LSTMs have also been proposed. For instance, GANs have been used for pedestrian trajectory prediction~\cite{social-gan} and CNNs have been used for traffic prediction~\cite{cnnpredict1}.

\textbf{Our Pedestrian Prediction:} We use a modified version of a state-of-the-art traffic trajectory prediction algorithm called RobustTP \cite{chandra2019robusttp} for pedestrian trajectory prediction. An image with the bounding box coordinates of the detected pedestrians (from YOLOv3) is fed  as input into RobustTP. This algorithm uses a hybrid structure with a combination of CNNs and LSTMs to predict the positions of the detected pedestrians in the next frame immediately after the input image. {The pedestrian tracking by YOLOv3 is shown in Fig.\ref{fig:TrackPredict} (a), and the position as predicted by RobustTP is shown in Fig.\ref{fig:TrackPredict}(b) as a black bounding box on a white background.} 

\textbf{Navigation using prediction}: Pedestrian prediction using Bayesian estimation and modeling their motions using RVO is presented in ~\cite{BRVO}. Long-term path prediction based on Bayesian learning and personality trait theory for socially-aware robot navigation is presented in \cite{Sociosense}.  Other techniques use a Partially Observable Markov Decision Process (POMDP) to model the uncertainties in the intentions of pedestrians. \cite{POMDP} presents a POMDP-based planner to estimate the pedestrians' goals for autonomous driving. The planner can be augmented with an ORCA-based pedestrian motion model~\cite{PORCA}. The  resulting POMDP planner runs in near real-time and can choose actions for the robot. Our approach is complimentary and can be combined with these navigation methods.

\subsection{Learning-Based Collision Avoidance with Pedestrians}
In recent years, several works have used different learning methods for navigation in dense scenes. GAIL (Generative Adversarial Imitation Learning) \cite{WB1} used raw depth camera images to train a socially acceptable navigation method for a differential drive robot. Similarly, \cite{End2End} used CNNs with RGB images to train an end-to-end visuomotor navigation system and \cite{D3QN} used a deep double-Q network (D3QN) to predict depth information from RGB images for static obstacle avoidance. \cite{CesarCadena} used demonstrations from an expert in simulation to train a method for mapless navigation. These methods, however, may not work well on dense crowds. 

A decentralized, scalable, sensor-level collision avoidance method was trained in \cite{JiaPan1}. It was extended to a hybrid learning architecture \cite{JiaPan2}, which switched policies based on the obstacle density in the environment. This method was further augmented to learn localization recovery points in the environment to solve the loss of localization and freezing robot problems simultaneously~\cite{Unfrozen}. Cooperative behaviors between humans and robots were modeled using a value network for better collision avoidance in ~\cite{JHow1}. \cite{JHow2} extended this formulation to observe an arbitrary number of pedestrians in the surroundings using LSTMs. \cite{Alahi} presented an approach to model interactions within a crowd which indirectly affect robot navigation. A novel collision avoidance method that identified previously unseen scenarios to carefully navigate around pedestrians is presented in \cite{JHow-uncertainty}.

\subsection{\textbf{DRL-based Collision Avoidance}}
Our learning-based algorithm is based on deep reinforcement learning.
The underlying objective is to train a policy $\pi_{\theta}$ that drives the robot to its goal while avoiding all the obstacles in the environment. 
There are three important components in DRL policy training, namely, \textbf{(i)} robot's observation space, \textbf{(ii)} action space, and \textbf{(iii)} the reward function. We briefly describe our observation and action spaces here and describe our novel reward function in Section III.

\textbf{Observation and Action Spaces:} Our observation space at any time instant \textit{t} can be represented as $\textbf{o}^t = [ \textbf{o}^t_{percep}, \textbf{o}^t_{odom}, \textbf{o}^t_{goal}]$, where $\textbf{o}^t_{percep}$ is the observation from the perception sensors, $\textbf{o}^t_{odom}$ is the odometry observation (which includes the current velocity of the robot), and $\textbf{o}^t_{goal}$ denotes the goal position relative to the robot. The action space of the robot is a continuous space consisting of the linear and angular velocities of the robot, represented as $\textbf{a}^t = [v^t, \omega^t]$.  




The robot performs a certain action until it receives the observation for the next time instant $\textbf{o}^{t+1}$. For optimizing the policy, we use the minimization of the mean arrival time of the robot to its goal as our objective function:

 \begin{equation}
     \underset{\pi_{\theta}}{\operatorname{argmin}}  \mathop{\mathbb{E}}[\frac{1}{N}\sum_{i = 1}^{N}t^g_i | \pi_{\theta}].
     \label{eqn1}
 \end{equation}

\subsubsection{\textbf{Proximal Policy Optimization}}
To train our collision avoidance policy, we use a policy gradient method \cite{sutton} called Proximal Policy Optimization (PPO) \cite{PPO}. Policy gradient methods (in contrast with value-based methods) directly modify the policy during training, which is more suitable for navigation applications and continuous action spaces. In addition, PPO bounds the update of parameters $\theta$ to a trust region \cite{trpo}, thereby ensuring that the policy does not diverge between two consecutive training iterations. This guarantees stability during the training phase.

%% file: Sections/4_OurMethod.tex
\section{Our Hybrid Approach: DenseCAvoid}
In this section, we present our hybrid  collision avoidance method that combines deep reinforcement learning with explicit pedestrian trajectory prediction for navigation. We present our network architecture that is used training our collision avoidance policy with anticipatory behaviors. We also discuss our reward function design and the complex 3-D training scenarios to handle dense crowds.




\subsection{Anticipating Pedestrian Behavior}
In densely crowded scenarios, pedestrian motion is highly unpredictable and non-smooth. In our case, we choose to explicitly model this behavior, similar to classic navigation methods. In particular, we use an explicit pedestrian trajectory predictor and use this information to generate non-oscillating, non-jerky navigation in dense scenarios. In addition, since our trained policy \textit{knows} where each pedestrian is headed in the immediate future, we can also make the robot avoid regions where several pedestrians might be heading. Therefore, the robot tends to avoid scenarios that could possibly lead to the freezing robot problem. 

As shown in Fig.\ref{fig:TrackPredict}, the prediction frame extracted using~\cite{chandra2019robusttp} has black bounding boxes at locations,  where the pedestrians could be in the future. These boxes are placed over a white background, which represents the collision-free free-space. We provide this \textit{future} free-space representation while training our network, which makes our collision avoidance policy training converge faster. This is due to the fact that our free-space representation provides a more direct way to infer the direction the robot should move towards and to learn the dynamic properties or behaviors of the pedestrians in different settings.

In our end-to-end formulation, we integrated the prediction with the DRL collision avoidance network as a new observation. Formally,  $\textbf{o}^t = [ \textbf{o}^t_{lid}, \textbf{o}^t_{cam}, \textbf{o}^t_g, \textbf{o}^t_v, \textbf{o}^t_{pred} ]$, where $\textbf{o}^t_{lid}$ denotes raw data from a 2-D lidar, $\textbf{o}^t_{cam}$ denotes the raw image data from either a depth or an RGB camera, $\textbf{o}^t_g$ refers to the relative position of the goal with respect to the robot, $\textbf{o}^t_v$ denotes the robot's current velocity, and $\textbf{o}^t_{pred}$ refers to the predicted positions of pedestrians in the next frame. 
Once the policy is trained, we sample a collision-free action from $\textbf{a}^t$ at each time instant as:
\begin{equation}
    \textbf{a}^t \sim \pi_{\theta}(\textbf{a}^t | \textbf{o}^t).
    \label{action}
\end{equation}
Note that it is also possible to directly combine the prediction output with the action $\textbf{a}^t$. This could be useful in scenarios where the robot encounters a situation that is quite different from the training data.


\subsection{Network Architecture}
\begin{figure}[t]
      \centering
      \includegraphics[height=2.0in,width=\columnwidth]{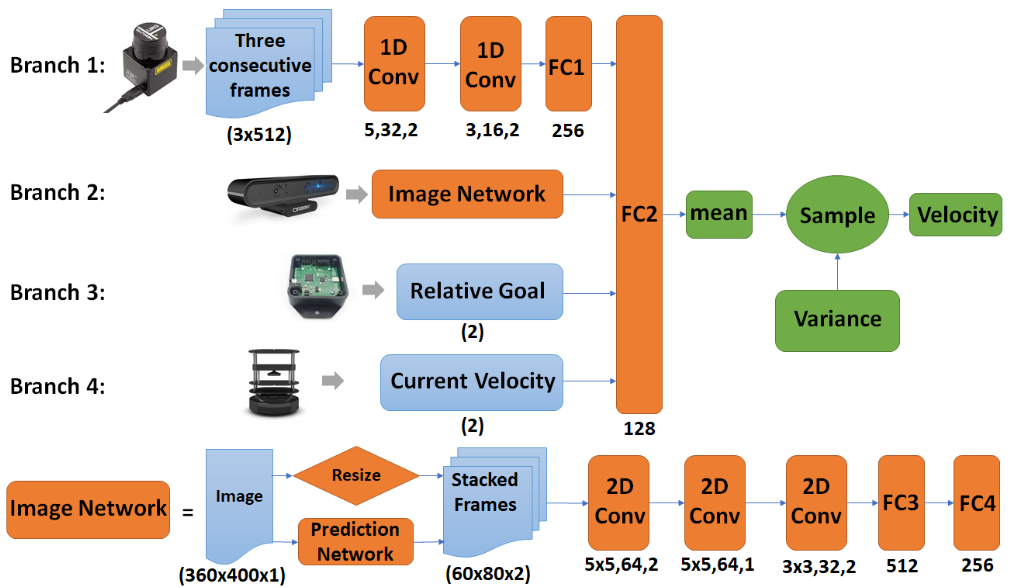}
      \caption {Architecture of our anticipatory collision avoidance network with four branches to process different observations. The input layer is marked in blue and, the hidden layers are marked in orange and the fully-connected layers in the network are marked as FCn. The green layer represents the output layer. The three values underneath each hidden layer denote the kernel size, number of filters, and stride length respectively. We stack a raw depth image with a the prediction frame and use it directly for training our collision avoidance policy.}
      \label{fig:Network}
      \vspace{-15pt}
   \end{figure}

Our network (Fig.\ref{fig:Network}) consists of four branches, each processing a component of the observation $o^t$. Two 1-D layers and three 2-D layers are used respectively for processing the 2-D lidar data and depth image data, which are followed by fully-connected layers that modify the dimensions of the outputs of the two branches to match each other. In branch 2, the depth image from the camera is first passed into our prediction algorithm. We then stack the computed prediction frame behind a resized version of the original depth image before passing these frames through a set of three 2-D convolutional layers. ReLU activation is applied to the outputs of all the hidden layers in branches 1 and 2. Branches 3 and 4 feed the relative position of the goal and the robot's current velocity to the fully-connected layer FC2. 

We apply a sigmoid activation to restrict the robot's linear velocity between (0.0, 1.0) m/s and a tanh activation to restrict the angular velocity between (-0.4, 0.4) rad/s in the output layer. The output velocity is sampled from a Gaussian distribution that uses the mean value outputted from the fully connected layer FC2, which is updated during training.

\subsection{Reward Function}
Our purpose during policy training is to reach the goal while avoiding collisions and to reduce oscillations or freezing behavior during navigation. Therefore, reaching the goal and colliding with obstacles are assigned high values of reward and penalty, respectively. To obtain smooth trajectories during run-time, we penalize sudden, large changes in the angular velocity. In addition, we use intermediate waypoints, which guide the policy away from obstacles and towards the goal, which results in faster convergence.

Formally, the total reward collected by a robot \textit{i} at time instant \textit{t} can be given as:
\begin{equation}
    \textit{r}_i^t = (r_g)^t_i \,+ (r_c)^t_i \,+ (r_{osc})^t_i \,+ (r_{safedist})^t_i,
\end{equation}
where the reward for reaching the goal $(r_g)^t_i$ or an intermediate waypoint is given as:

\begin{figure*}[t]
\includegraphics[height=1.7in, width=\linewidth]{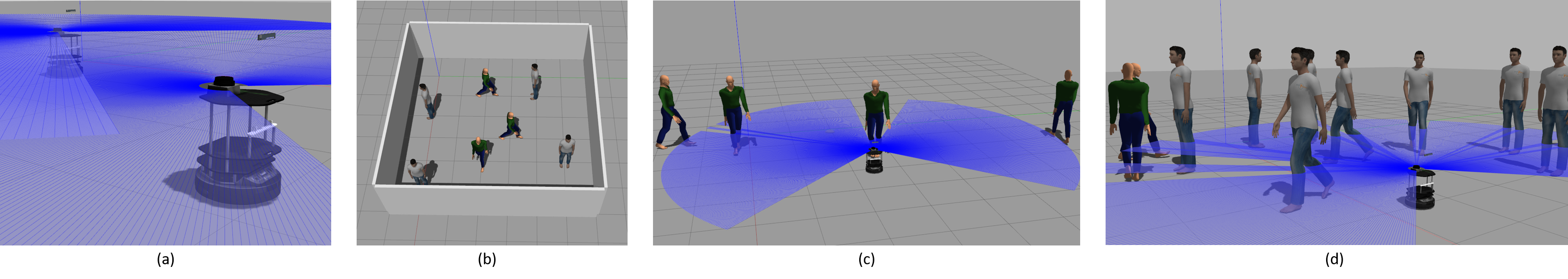}
\caption{Different training scenarios used for training our algorithm from simplest to complex. \textbf{(a)}: Empty scenario with random goals; \textbf{(b)}: Dense-Static scenario with random goal; \textbf{(c)}: Robot moves through a few pedestrians walking randomly to reach the goal; \textbf{(d)} Robot moves through a dense crowd to reach its goal. }
\label{fig:reward}
\vspace{-5pt}
\end{figure*}

\begin{equation}
    (r_g)^t_i =
    \begin{cases}
     r_{wp} \qquad \qquad \qquad \qquad if \, ||\textbf{p}_i^t - \textbf{p}_{wp}|| < 0.2,\\
     r_{goal} \qquad \qquad \qquad \qquad if \, ||\textbf{p}_i^t - \textbf{g}_i|| < 0.3,\\
     2.5(||p_i^{t-1} - g_i|| - ||p_i^t - g_i||) \qquad   otherwise.
    \end{cases}
\end{equation}
The collision penalty $(r_c)^t_i$ is given as: 
\begin{equation}
    (r_c)^t_i =
    \begin{cases}
     r_{collision} & if \, ||\textbf{p}_i^t - \textbf{p}_{obs}|| < 0.3,\\
     0  &  otherwise.
    \end{cases}
\end{equation}
The oscillatory behaviors (i.e. choosing sudden large angular velocities) are penalized as:
\begin{equation}
    (r_{osc})^t_i = -0.1 |\omega_i^t| \qquad \qquad  if \, |\omega_i^t| > 0.3.
\end{equation}
The penalty for moving too close to an obstacle is given by:
\begin{equation}
    (r_{safedist})^t_i = -0.1 ||R_{S_{max}} - R_{min}^t||.
\end{equation}
We set $r_{wp}$ = 10, $r_{goal}$ = 20, and $r_{collision}$ = -20 in our formulation. 


\subsection{Training Scenarios}

The policy training is carried out in multiple stages to ensure fast convergence of the total accumulated reward. We designed several training scenarios to suit our pedestrian prediction network by including walking pedestrian models in the dynamic scenes in our training.  Furthermore, we include dense pedestrian scenarios with sudden changes in the path. Our training scenarios are designed with these parameters:

\begin{itemize}
    \item \textbf{Random Goal:} The robot is given a random goal in an empty world, and actions leading the robot towards the goal are rewarded. In this scenario, the partially trained model learns basic goal-reaching capabilities and is saved for training in the next scenario.
    
    \item \textbf{Dense-Static:} The robot is given a random goal in a world cluttered with static obstacles. During training, the policy augments its previously learned goal-reaching capabilities with basic static obstacle avoidance.
    
    \item \textbf{Random-Pedestrians:} The policy from the previous scenario is now trained in a world with randomly walking pedestrians. The pedestrian prediction observations now play a major role in training the policy for dynamic obstacle avoidance. We can varying the position, trajectories, and the densities of the pedestrians.
    
    \item \textbf{Dense-Random-Pedestrians:} In this scenario, the robot needs to navigate through a dense crowd of randomly walking pedestrians before reaching its goal. 
    
\end{itemize}

\begin{figure}[t]
\includegraphics[height=1.8in, width=\columnwidth]{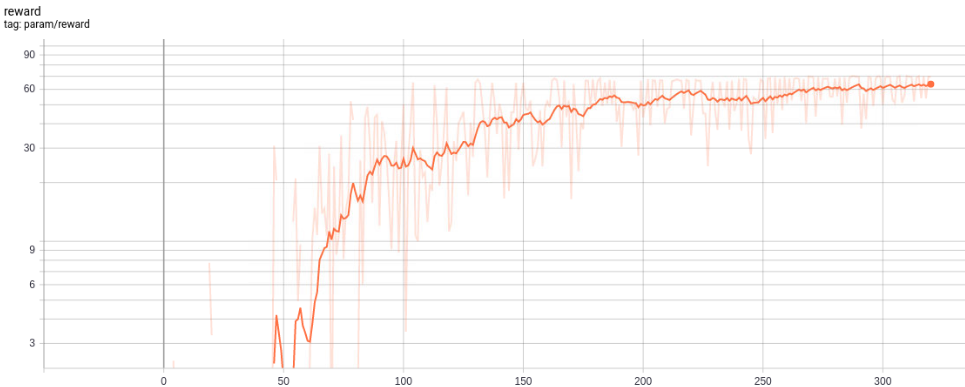}
\caption{Convergence of our reward function vs the number of iterations. Explicit pedestrian prediction helps our training converge faster and smoother when compared with the case with no prediction.}
\label{fig:reward3}
\vspace{-10pt}
\end{figure}

%% file: Sections/5_Results_and_Evaluations.tex
\begin{table*}[ht]
\resizebox{\linewidth}{!}{
\begin{tabular}{|c|c|c|c|c|c|} 
\hline
Metrics & Sensor Configuration & Dense-Static & Random-Sparse-Ped & Dense-Ped \\ [0.5ex] 
\hline
\multirow{3}{*}{Success Rate} & Depth Camera  & 0.26 & 0.73 & 0.55   \\
 & Depth Camera + Lidar & 0.6 & 0.733 & 1  \\
 & {DenseCAvoid} & 0.93 & 0.87 & 1 \\
\hline

\multirow{3}{*}{Avg Trajectory Length} & Depth Camera  & N/A & 5.5 & 16.3 \\
 & Depth Camera + Lidar & N/A & 5.11 & 15.51  \\
 & {DenseCAvoid} & N/A & 6.39 & 16.87 \\
\hline


\end{tabular}
}
\caption{\label{tab:results} Relative performance of our DenseCAvoid hybrid method versus learning-based methods that do not use explicit pedestrian prediction. In the latter category, we use two combinations of sensors (i.e. only depth camera and depth camera + lidar). The trajectory length for Dense-Static case is not measured, since we assign the robot's goals randomly.  These numbers highlight the benefit in terms of success rate (higher is better) of DenseCAvoid. However, the explicit use of prediction can slightly increase the trajectory length (lower is better) to the goal, because the robot may not take the shortest straight line path. 
}
\end{table*}




\begin{table*}[h]
\resizebox{\textwidth}{!}{
\begin{tabular}{|c|c|c|c|c|} 
\hline
Metrics  & Distance & Depth Camera & Depth camera + lidar & DenseCAvoid \\ [0.5ex] 
\hline
\multirow{3}{*}{Freezing Robot \%} & $<$ 1.0 meters  & 100\% & 100\% & 100\%   \\
 & 1 - 1.5 meters & 53\% & 33\% & 5\%  \\
 & 1.5 - 2 meters & 27\% & 0\% & 0\% \\
\hline

\end{tabular}
}
\caption{\label{tab:freezing} The performance of different methods in avoiding freezing robot scenarios, i.e. the number of instance the robot freezes (lower is better), tested in the Robot Freezing scenario. Our DenseCAvoid method considerably improves the performance when the robot is more than $1$m away from an obstacle. In these cases, our explicit prediction of pedestrian trajectory improves the navigation capability and the robot does not freeze.}
\end{table*}

\section{Results and Evaluations}
In this section, we describe our implementation and highlight the performance of DenseCAvoid in different scenarios. We also compare our navigation algorithm with policies trained without trajectory prediction and highlight the benefits of explicitly modeling the anticipatory pedestrian behavior. The convergence of the logarithm of our reward function versus the number of iterations is shown in Fig. \ref{fig:reward}.

\subsection{Implementation}
Our policy is trained in simulations created using ROS Kinetic and Gazebo 8.6 on a workstation with an Intel Xeon 3.6GHz processor and an Nvidia GeForce RTX 2080Ti GPU. We use Tensorflow, Keras, and Tensorlayer to create our network. 
We simulate sensor data using models of the Hokuyo 2-D lidar and the Orbbec Astra depth camera in Gazebo during training and testing. The 2-D Hokuyo lidar has a maximum range of $4$ meters, has a field of vision (FOV) of 240$^\circ$, and provides $512$ range values per scan. The Orbbec Astra has a minimum range of $1.4$ meters and a maximum sensing range of $5$ meters. We use depth images of size $60 \times 80$ as inputs to our policy training network. We use a low resolution image to reduce the latency for processing the data. For pedestrian prediction, we normalise the depth images before passing it into  a pre-trained YOLOv3 and RobustTP network. We observe about $85$\% prediction accuracy in our dense benchmarks. 
We mount the same sensors on a Turtlebot 2 robot to test our model in real-world scenarios such as densely crowded corridors with non-smooth pedestrian trajectories.


\subsection{Testing scenarios}
We compare our policy trained with pedestrian prediction with two policies which were trained without pedestrian prediction: (i) Policy trained only with depth camera observations, and (ii) Policy trained with lidar and depth camera observations. We consider four different test scenarios that have more challenging static and dynamic scenes, as compared to our training scenarios. This demands tight maneuvers from the robot to reach its goal. The scenarios we consider are:

1. \textbf{Dense-Static}: Scenario cluttered with static obstacles, with random goals provided to the policy.

2. \textbf{Random-Sparse-Ped}: Scenario where the robot has to pass through $10$ randomly walking pedestrians to reach its goal. 

3. \textbf{Dense-Ped}: Scenario where the robot must move against the direction of $15$ walking pedestrians in a narrow corridor to reach its goal.   

4. \textbf{Robot Freezing}: Several ($3-4$) pedestrians are suddenly spawned at different distances ($1-2$ meters) in front of the robot to simulate the freezing-robot scenario. Most prior benchmarks fail in such cases and we highlight the benefits of explicit trajectory prediction.   

\subsection{Performance Benchmarks and Metrics}
We use the following  metrics to evaluate the performance of different navigation algorithms:
\begin{itemize}
\item \textbf{Success Rate} - The number of times that the robot arrived at its goal without colliding with an obstacle over the total number of attempts.

\item \textbf{Average Trajectory Length} - The trajectory length that the robot travels before reaching the goal, calculated as the sum of linear segments over small time intervals over the total number of attempts.


\item \textbf{Robot Freezing \%} - The number of times the robot got stuck or started oscillating indefinitely, while avoiding sudden obstacles over total number of attempts. A lower value is better.
\end{itemize}

\subsection{Analysis and Comparison}
We present our results in Table \ref{tab:results}. We observe that DenseCAvoid consistently has a higher success rate as the testing scenarios get more complicated. Using only observations from the depth camera  for navigation works well for sparse scenarios, however, performs poorly in dense scenarios. This is mainly due to the low field of vision of the depth camera when compared to a lidar. Using observations from depth camera and a 2-D lidar performs slightly better than a single sensor. However, explicit modeling of the pedestrian future trajectory and behavior results in improvement in these scenarios. However, the use of trajectory prediction can increase the trajectory length, as the robot may take a larger turn during collision avoidance. 

The main benefit of DenseCAvoid arises in terms of dealing with the freezing robot problem, as can be observed from Table \ref{tab:freezing}. All methods fail in scenarios where a pedestrian suddenly appears within 1 meters from the robot.  This is a consequence of limiting the angular velocity of the robot between (-0.4, 0.4) rad/sec to avoid high oscillations during navigation. Prior deep reinforcement learning methods can't deal with such situations and the robots tend to freeze. However, DenseCAvoid is able to handle sudden pedestrians about $1$ to $1.5$ meters away much better than the other methods, leading to oscillations/freezing only 5\% of the time. This highlights the benefit of using explicit pedestrian prediction as the classic navigation method along with learning-based methods.


%% file: Sections/6_Conclusion.tex
\section{CONCLUSIONS, LIMITATIONS, FUTURE WORK}
We present a novel method for collision avoidance with pedestrians in dense scenarios. Our hybrid approach tends to combine the benefits of learning-based methods with classic navigation methods that explicitly perform trajectory prediction.  Our approach has been implemented in highly dense crowds with pedestrian densities up to 1-2 pedestrian per square meter. Our prediction method does not make any assumptions regarding the motion model for the pedestrian, which results in stable behavior during training. We validate our work in simulation and in real-world scenarios using a Turtlebot. We showed that our approach drastically reduces the freezing robot problem, when pedestrians suddenly appear in front of the robot. Our work has several limitations. While we improve the navigation capabilities in dense scenarios, our approach does not work robustly in all possible scenarios. The accuracy is also limited by the accuracy of trajectory prediction, which is not perfect. The overall performance is mostly governed by the synthetic datasets used during the training phase and the limitations of sim-to-real paradigm. As part of future work, we want to overcome these limitations. It may also be possible to combine classic navigation methods as a post-process to the action computed by Equation \ref{action}. We also need to take into dynamics constraints of the robot and test them in outdoor scenarios. We also need better perception techniques to handle transparent surfaces.
